\documentclass[runningheads]{llncs}

\usepackage{eccv}

\usepackage{eccvabbrv}

\usepackage{graphicx}
\usepackage{booktabs}
\usepackage{algorithm}
\usepackage{algorithmic}

\makeatletter
\newcommand\notsotiny{\@setfontsize\notsotiny\@vipt\@viipt}
\makeatother

\usepackage[accsupp]{axessibility}  % Improves PDF readability for those with disabilities.

\usepackage{hyperref}

\usepackage{orcidlink}

\begin{document}

\title{Dataset Distillation by Automatic Training Trajectories} 

% \titlerunning{Dataset Distillation by Automatic Training Trajectories}

\author{Dai Liu\inst{1}\orcidlink{0009-0007-2530-4595} \and
Jindong Gu\inst{2}\orcidlink{0009-0000-0574-0129} \and
Hu Cao\inst{1}\orcidlink{0000-0001-8225-858X} \and
Carsten Trinitis\inst{1}\orcidlink{0000-0002-6750-3652} \and
Martin Schulz\inst{1}\orcidlink{0000-0001-9013-435X}}

\authorrunning{D. Liu et al.}

\institute{Technical University of Munich, Germany \and
University of Oxford, United Kingdom}

\maketitle
% {\let\thefootnote\relax\footnotetext{*: Corresponding author}}
\begin{abstract}
% 70-150 words
Dataset Distillation is used to create a concise, yet informative, synthetic dataset that can replace the original dataset for training purposes. Some leading methods in this domain prioritize long-range matching, involving the unrolling of training trajectories with a fixed number of steps ($N_{S}$) on the synthetic dataset to align with various expert training trajectories. However, traditional long-range matching methods possess an overfitting-like problem, the fixed step size $N_{S}$ forces synthetic dataset to distortedly conform seen expert training trajectories, resulting in a loss of generality—especially to those from unencountered architecture. We refer to this as the Accumulated Mismatching Problem (AMP), and propose a new approach, Automatic Training Trajectories (ATT), which dynamically and adaptively adjusts trajectory length $N_{S}$ to address the AMP. Our method outperforms existing methods particularly in tests involving cross-architectures. Moreover, owing to its adaptive nature, it exhibits enhanced stability in the face of parameter variations. Our source code is publicly available at \url{https://github.com/NiaLiu/ATT}
\end{abstract}
\begin{figure}[t]
    \centering
    % \subfloat[The plot demonstarte\label{fig:testb}]
  {\includegraphics[width=.3\linewidth]{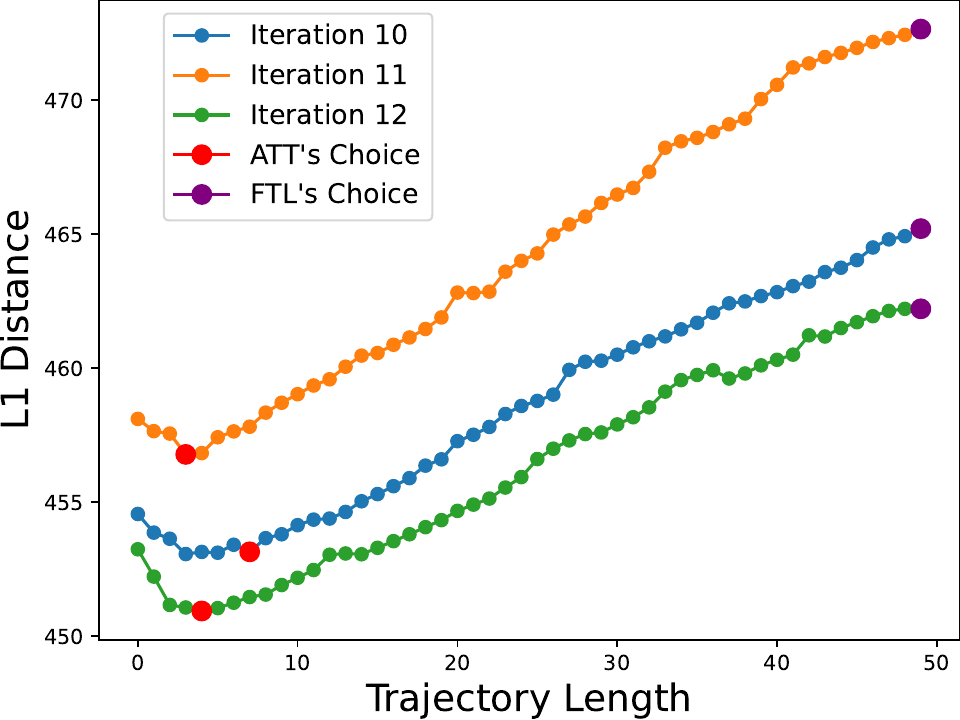}}\hspace{10pt}
  % \subfloat[ATT avoid mismatching when by-passing the target at higher iterations\label{fig:testc}]
  {\includegraphics[width=.3\linewidth]{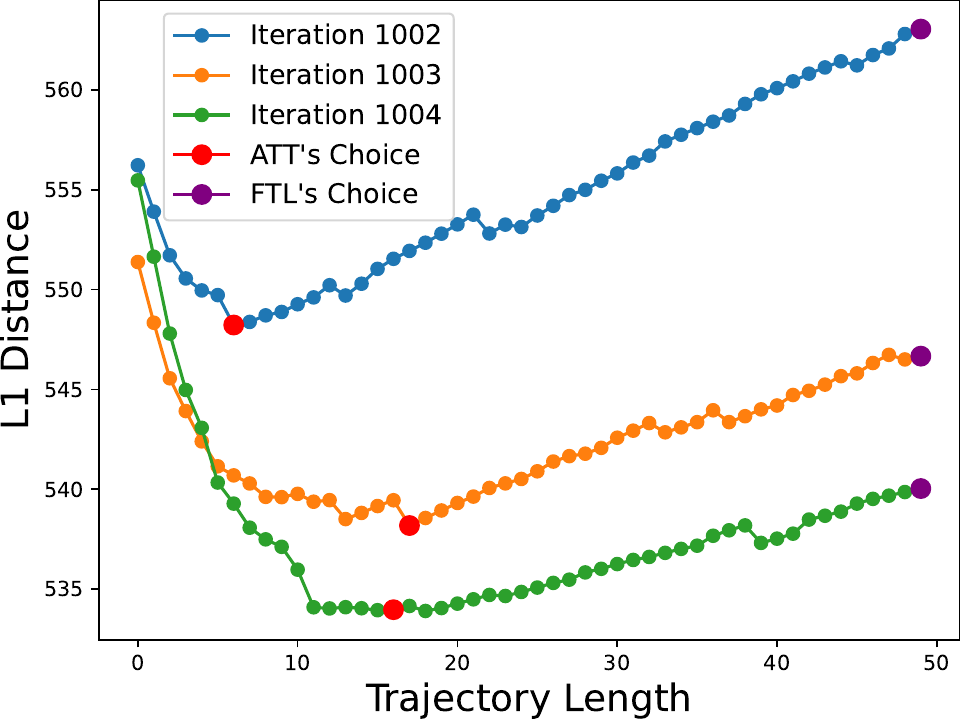}}
  \caption{\label{fig:demo} The plots shows L1 distance between each network from a training trajectory and the corresponding target, and chosen target by different method. The experiments are carried on CIFAR-10. Existing methods employs Fixed Training Length (FTL), which select network at the end of a trajectory. But our method ATT dynamically selects network possessing closest distance to targets. Left plot shows examples from beginning iterations, and right plot shows higher iterations. ATT dynamically adjusts matching target, thus avoid large matching error and unwanted stretching of trajectories.}
\end{figure}

\section{Introduction}
\label{sec:intro}
Deep learning has showcased remarkable achievements across various computer vision problems~\cite{dlcv-1, dlcv-2, dlcv-3}. Nevertheless, its success often hinges on extensive datasets, resulting in substantial computational demands. Recognizing the escalating computational costs~\cite{greenai, gpuera}, numerous academic publications have delved into addressing these challenges. A key focus has been on mitigating the intensive training process for end-users by developing significantly reduced datasets. A conventional method for dataset reduction is Coreset Selection (CS)~\cite{cs1, cs2, cs3, cs5, cs6}, which involves curating a subset of the most informative training samples. However, many CS methods are characterized as greedy algorithms, leading to a trade-off between speed and accuracy~\cite{cs-herding, cs-forgetting, cs1, cs-greedy1, cs-greedy2}. Additionally, the information provided by these methods is constrained by the selected samples from the original dataset. In 2018, Wang et al.~\cite{dd} introduced Dataset Distillation (DD), a technique that prioritizes the extraction of an informative and compact dataset capable of replacing the original training dataset in various training tasks. The technique not only provides a significantly smaller synthetic dataset for training but also conceals details in the original dataset, thereby enhancing privacy~\cite{dd-privacy}. DD methods have been widely applied in various fields, including federated learning~\cite{dd-fed0, dd-fed1, dd-fed2, dd-fed-3, dd-fed-4}, continual learning~\cite{dd-continue1, dd-continue2, dd}, studies on attacks~\cite{dd-attacks1, dd-attacks2,dd}, and many other areas~\cite{dd-am, dd-MTT-fasion, dd-decentrilise-learning, dd-apply-graphs, dd-apply-medical}.

% Introduce DD history 
The DD methodologies can be classified into two distinct categories based on their alignment targets: short-range matching methods~\cite{dd-dc, dd-dsa, dd-cafe, dd-kip, dd-dm} and long-range matching methods~\cite{dd-mtt, dd-prunning, dd-mae}. Short-range matching strategies, notably initiated by a gradient-matching approach~\cite{dd-dc}, concentrate on aligning a single training step executed on distilled data with that performed on the original data. Conversely, long-range matching Dataset Distillation (LDD) entail the alignment of multiple training steps. Empirical validation by Cazenavette et al.~\cite{dd-mtt} demonstrates that long-range matching consumes more computation but exhibits superior performance in terms of test outcomes comparably, and attributed the great test performance to the ability to circumvent short-sightedness. Following research improved LDD further, such as with flat trajectories~\cite{dd-mae}, with prunning to filter out the hidden noise~\cite{dd-prunning}, or investigating training stages~\cite{dd-datm}.

Nonetheless, our empirical findings expose a common trend in conventional LDD methods, where inaccurate predictions are consistently reinforced throughout the majority of iterations. We attribute this issue to their reliance on Fixed Trajectory Length (FTL), which lacks the adaptability to align various steps of expert trajectories, as shown in~\cref{fig:demo}. This reinforcement, driven by the FTL, results in the accumulation of matching errors over the distillation process. Importantly, these errors not only persist, but accumulate in the synthetic dataset. Consequently, the synthetic dataset tends to overly conform to observed matching targets while struggling to generalize effectively to unseen matching targets. We term this issue the AMP (Accumulated Mismatching Problem).

Our studies underscore the significance of this observation, and we propose ATT (Automatic Training Trajectories). ATT is specifically designed to dynamically and adaptively adjust matching objects and trajectory length, aiming to diminish AMP throughout the distillation process. Our method paves the way towards adaptive and dynamical matching, fostering improved generalization and accuracy in synthetic dataset distillation.

In summary, our main contributions are as follows:
\begin{itemize}
\item We revisit the domain of long-range matching dataset distillation, and identify the Accumulated Mismatching Problem (AMP).
\item We illustrate how AMP contributes to the iterative matching error and establish a clear correlation between AMP and Accumulated Trajectory Error (ATE).
\item We present an innovative Long-range Matching Dataset Distillation (LDD) method named Automatic Training Trajectories (ATT). ATT incorporates a dynamic and adaptive approach to the selection of matching objects by adjusting the matching length, leveraging a minimum distance strategy to effectively address the challenges posed by AMP.
\end{itemize}

\section{Related Work}
\subsection{Dataset Distillation}
Dataset Distillation (DD) was initially introduced by Wang et al.\cite{dd}, drawing inspiration from Knowledge Distillation~\cite{KD}. 
% The general approach of DD involves expanding the latent knowledge of the dataset, including gradients, outputs, and network parameters. This expanded information is then matched to distill a smaller synthetic dataset. 
Existing DD baselines can be categorized into short-range and long-range matching methods based on the number of steps unrolled on the original dataset. In this context, kernel ridge regression-based methods are included under short-range matching.\vspace{-5pt}\newline\\
{\bf Short-Range Matching Dataset Distillation (SDD):} Dataset Condensation (DC), introduced by Zhao et al.~\cite{dd-dc}, has been a subject of extensive research and it involves matching one-step updated gradients. Subsequent research by Zhao et al.~\cite{dd-dm} enhanced performance and reduced synthesis costs by incorporating Batch Normalization and aligning feature distributions. \cite{dd-idm},\cite{dd-dm} further addressed the problem of miss-alignment among classes. \cite{dd-dcc} focuses on capturing differences between classes, and \cite{dd-cafe} overcomes the bias of gradient matching methods by aligning layer-wise features. \cite{dd-dsa} enhanced DC by boosting the distillation process with Differentiable Siamese Augmentation (DSA). \cite{dd-squeeze} presents a new DC framework decouples bi-level optimization, and makes a step towards real world applications. Approaches applying a kernel-based meta-learning framework with an infinitely wide neural network~\cite{dd-kip} have also sparked significant research. This includes methods implementing model pool and label learning~\cite{dd-frepo}, and adopting convexified implicit gradients~\cite{dd-cig}.\vspace{-5pt}\newline\\
{\bf Long-Range Matching Dataset Distillation (LDD):~\label{subsubsec:LDD}} DD, as introduced by \cite{dd}, initially focused on one-step matching. The concept of Matching Training Trajectories (MTT) was later introduced by \cite{dd-mtt} to replicate the long-range training dynamics of the original dataset. This approach overcomes the short-sightedness in single-step distillation and introduced the first long-range matching method, referred to as the vanilla LDD method in this work. Li et al.~\cite{dd-prunning} implements Parameter Pruning (PP) to enforce matching on significant parameters. Additionally, \cite{dd-mae} discovers the Accumulated Trajectory Error (ATE) in the distillation and evaluation phases, adopting flat expert trajectories to alleviate ATE, and~\cite{dd-datm} reveals different training stages contains different information.\vspace{-5pt}\newline\\
{\bf Divergent Research Focus: }
Other concurrent works have improved DD baselines with various approaches. Some have adopted soft labels~\cite{dd-sl, dd-sl1, dd-tasla}, while others have implemented data parameterization~\cite{dd-idc, dd-am, dd-slimmable} and data factorization~\cite{dd-dcfactor, dd-haba, dd-prior, dd-gan}. Another approach involves encoding labels and synthetic datasets into a learnable addressing matrix~\cite{dd-am}. Some works focus on selecting representative real samples~\cite{dd-ddrame}, using model augmentation~\cite{dd-model-aug}, and reducing the budget for storage and transmissions with slimmable DD~\cite{dd-slimmable}. Additionally, there are efforts focusing on using only a few networks~\cite{dd-fewshot}.

\subsection{Sample Selection}
Like DD, instance selection~\cite{cs-is}, coreset selection~\cite{cs1, cs2, cs5}, and dataset pruning~\cite{cs3} also aim to identify representatives from a given dataset. These sample selection methods have been widely applied in diverse fields~\cite{cs-u1, cs-u3, cs-u4, cs-u5, cs-u2}. Established techniques, such as random selection, herding methods~\cite{cs-herding, cs-herding1}, and forgetting~\cite{cs-forgetting, cs-forgetting1}, can be adapted as comparisons to dataset distillation methods. While similar to other selection methods, DD, in its aim of creating information-rich images that may not necessarily be considered actual samples. Multiple studies~\cite{dd-dc, dd-dsa, dd-mtt} have empirically proven that dataset distillation, especially at higher compression ratio, significantly outperforms sample selection.
\section{Method}
Conventional LDD methods typically unroll a fixed length of trajectory on the synthetic dataset to match expert trajectories. However, a fixed trajectory length cannot effectively handle variations within expert trajectories, leading to mismatched predictions and accumulated errors in the synthetic dataset, which we refer to the Accumulated Mismatching Problem (AMP). In this section, we first introduce the conventional LDD and the typical matching strategy involving Fixed Trajectory Length (FTL) in \cref{subsec:FTL}. Subsequently, we empirically demonstrate the AMP in conventional LDD and its impact on Accumulated Trajectory Error (ATE) in \cref{subsec:bottleneck}. Lastly, we present our proposed method, Automatic Training Trajectories (ATT), designed to address AMP, and provide the pseudo-code in \cref{subsec:ATT}.

\subsection{LDD}
\label{subsec:FTL}
{\bf Problem Description:} LDD methods reveal the information within the real dataset $D_{T}$ through the creation of training trajectories. Specifically, we define a training trajectory as a function $T_{D,f}(\theta_0, N)$. The function utilizes classifier $f$ initialized with network parameters $\theta_0$ and optimizes the classifier iteratively for a total of $n$ steps over a dataset $D$. $T_{D,f}(\theta_0, N)$ outputs the difference between the network parameters at the $N$-th iteration and those at the initial iteration. Namely, 
\begin{equation} \label{eq:1}
T_{D,f}(\theta_0, N) = \theta_N - \theta_0.
\end{equation}
LDD endeavors to enhance the synthetic dataset $D_{S}$ by aligning it with the expanded information in the form of trajectories. Specifically, the objective is to address the following optimization problem, as shown in the following:
\begin{equation} \label{eq:2}
\Delta T = \|T_{D_{S,f}}(\theta_{i}, N_{S})-T_{D_{T},f}(\theta_{i}, N_{T})\|^{2}_{2},
\end{equation}
\begin{equation} \label{eq:3}
D_{S} = \underset{D_{S}}{\arg \min{ }} \underset{\theta_{0}\sim P_{\theta_0}}{\mathbb{E}}[\Delta T].
\end{equation}
We use subscripts $_S$ and $_T$ to denote parameters associated with synthetic data and real data, respectively. $P_{\theta_0}$ is the probability distribution of network initialization.

In practical scenarios, the training trajectory is unrolled on the real dataset $D_{T}$, generating expert trajectories sets as $\{\theta^{*}_{0,0}, \theta^{*}_{0,1}, \cdots, \theta^{*}_{0,M}, \theta^{*}_{1,0},\cdots,\theta^{*}_{N,M}\}$. N is the number of trajectories, M is trajectory length. The expert set is pre-collected and stored in buffers, with additional disk storage to alleviate memory constraints.

Following the collection of the buffer, we initialize the student trajectory using any expert from the set of expert trajectories, such as $\theta^{\prime}_{i,0}={\theta^{*}_{i,0}}$. Subsequently, a similar training trajectory is unfolded on the synthetic dataset $D_{S}$ for $t$ steps, represented as ${\theta^{\prime}_{i,t}}$. In conventional LDD methods, a fixed value $N_S$ is chosen for $t$. The objective is to minimize the discrepancy between the student prediction ${\theta^{\prime}_{i,t}}$ and the expert trajectory unrolled at $N_T$ steps, denoted as $\theta^{*}_{i,N{T}}$. The optimization of $D_{S}$ is then carried out by minimizing the defined loss function
\begin{equation} \label{eq:4}
L = \frac{\|\theta^{\prime}_{i,N_{S}}-\theta^{*}_{i,N_{T}}\|^{2}_{2}}{\|\theta^{*}_{i,0}-\theta^{*}_{i,N_{T}}\|^{2}_{2}}.
\end{equation}
This loss function quantifies the distance between the final state of the student trajectory and the expert trajectory using the squared L2 norm. Subsequently, the distance is normalized by the squared Euclidean distance between the initial and final states of the expert trajectory, allowing for the amplification of weak signals in the expert sets.\newline\\
{\bf Fixed Trajectory Length}
LDD methods conventionally adopt a Fixed Trajectory Length (FTL) strategy, wherein a predetermined step size denoted as $N_S$ is fixed for all instances of trajectory matching. This fixed step size is uniformly applied to match various experts from the expert trajectory set. Previous studies, as observed in studies such as~\cite{dd-mtt, dd-mae}, often conducted experiments involving an exhaustive search to determine an optimal $N_S$. However, our investigation reveals that a fixed trajectory length introduces the Accumulated Mismatching Problem (AMP). This problem leads to persistent matching errors that do not diminish, even with an optimal $N_S$. The issue persists across variations in $N_S$ and remains throughout the entire distillation process. Further details are discussed in the subsequent~\cref{subsec:bottleneck}.

\begin{figure}[tb]
\centering
    \includegraphics[width=0.95\linewidth]{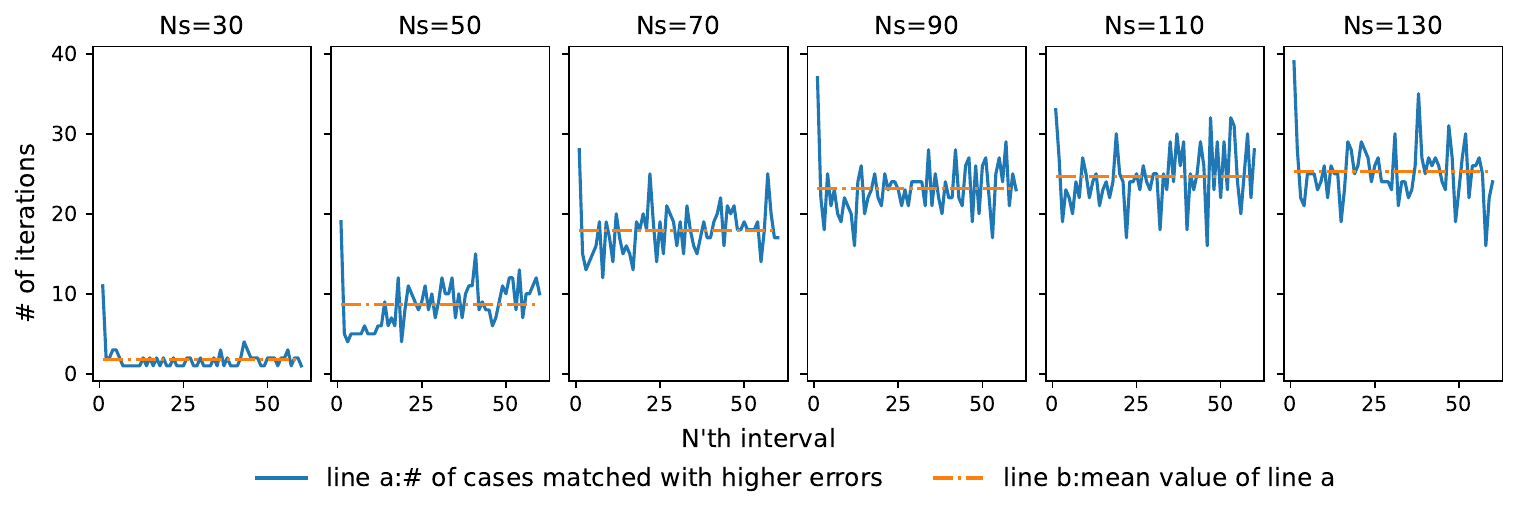}
   \caption{\label{fig:1a} The figure illustrates that Fixed Trajectory Length (FTL) matches all experts with avoidable matching error throughout the distillation process. The figure is generated from experiments conducted on CIFAR-10 with Images Per Class (IPC) set to 1. We collect the number of cases matches with larger matching errors $\|N_S-N_{opt}\|\geq\gamma$, at every 50 iterations throughout the distillation process. The number of cases matched with larger errors fluctuates over entire process. The same can be observed with the mean value of line a. From left to right, we observe the persistence of this issue across different step size $N_S$ for FTL. Notation: \#: number, N'th interval: the N'th 50 iters.
  }
\end{figure}
\subsection{Accumulated Mismatching Problem (AMP)\label{subsec:bottleneck}} 
In traditional LDD methods, a specific trajectory length denoted as $N_S$ is fixed while matching various segments of expert trajectories. Network parameters are predicted for all instances of matching based on this FTL, namely, $\theta^{\prime}_{i,N_S} = \theta^{\prime}_{i,0} + T_{D_S,f}(\theta^{\prime}_{i,0}, N_S)$. However, due to variations in expert trajectories arising from different steps on trajectories and different initialization of them, the designated prediction $\theta^{\prime}_{i,N_S}$ lacks the necessary adjustments to accommodate such variations among different experts. This discrepancy leads to the so-called Accumulated Mismatching Problem (AMP).

We provide details in experiments conducted on CIFAR-10 with a configuration of 1 Image Per Class (IPC). The chosen baseline is the vanilla LDD method~\cite{dd-mtt}, to mitigate any unintended buffered effects other than FTL. In our experimental analysis, we initiate the process by generating an exhaustive set of potential predictions $pred=\{\theta^{\prime}_{i,1}, \theta^{\prime}_{i,2},\cdots\theta^{\prime}_{i,N_S}\}$ under diverse values of $N_S$ ranging from 30 to 130.

Subsequently, we measure the mean square error between each prediction and the corresponding target. The $pred$ associated with the minimal error is identified at $\theta^{\prime}_{i,N_{opt}}$, at the optimal step $N_{opt}$. The existance of the gap between $N_{opt}$ and $N_S$ introduces extra unwanted matching error. To account for minor discrepancies, we introduce a parameter $\gamma=2$ as a tolerance threshold for errors. We then collect the number of cases where matches with larger error, specifically $\|N_S-N_{opt}\|_{1}\geq\gamma$. This collection is performed at each 50 iterations throughout the distillation process, and the results are depicted in \cref{fig:1a}. Additionally, we present supplementary results with varying $\gamma$ in \cref{appendix:gamma}.

As observed in the figure, FTL methods consistently show a tendency to align steps with larger errors across different values of $N_S$. This behavior indicates that FTL methods compel trajectory matching with an inappropriate trajectory length. Consequently, student trajectories with an inappropriate length undergo either compression or stretching to conform to the diverse patterns exhibited by expert trajectories. This compels the synthetic dataset to overfit to the pre-buffered expert trajectories but sacrifices generality for unseen expert trajectories. We validate this observation with experimental results in~\cref{subsec:Cross}, where AMP has been addressed. Additionally, the results obtained in~\cref{subsec:Cross} support that the challenge becomes particularly pronounced when synthetic data encounters unseen trajectories from unseen architectures.\vspace{-5pt}\newline\\
{\bf AMP is Accumulative: }
We investigate into $N_S$ by mimicing it as a small fluctuation $\delta$. An additional term emerges in the prediction, denoted as $\theta_{N_S+\delta} = \theta_{N_S} + T_{D_S,f}(\theta_{N_S},\delta)$. The presence of the additional term has implications for both the loss function, as depicted in~\cref{eq:4}, and the objectives outlined in~\cref{eq:2} and \cref{eq:3}. Consequently, this additional term accumulates with each update to the synthetic dataset $D_S$, a phenomenon we identify as accumulative.\vspace{-5pt}\newline\\
{\bf AMP and Accumulated Trajectory Error: }
Previous research~\cite{dd-mae} revealed the existence of Accumulated Trajectory Error (ATE), which results from discrepancies between the initialization from previous student parameters and the assigned initial expert parameters in each iteration. Conversely, while AMP primarily focuses on the end of the trajectory for matching, it is noteworthy that AMP also contributes to ATE. In this subsection, we show the correlation between AMP and ATE.

As outlined in from~\cite{dd-mae} Sec. 3.1, the ATE is composed of the initialization error, the matching error $\epsilon_{0}$, and the ATE from the previous iteration. The matching error $\epsilon_{0}$, as part of ATE, is defined as
\begin{equation}\label{eq:7}
    \epsilon_{0} = T_{D_S,f}(\theta^{*}_{i,0}, N_S)-T_{D_T,f}(\theta^{*}_{i,0}, N_T).
\end{equation}
$\theta^{*}_{i,0}$ is any network parameters from expert trajectories. We plug~\cref{eq:1} in~\cref{eq:7} and take L2 norm as distance measure. We define $e_{t}=\|\epsilon_{0}\|_2^2$, and AMP is indeed solving the following in each iteration as shown:
\begin{equation} \label{eq:8}
    \min_{t} e_{t} = \min_{t\in\{0,1,\cdots,N_S\}} \|\theta^{\prime}_{i,t}-\theta_{i,N_T}^*\|_2^2.
\end{equation}
By addressing the AMP, we minimize matching errors at each step, thus iteratively reducing the ATE.

\begin{figure*}[tb]
\centering
    \includegraphics[width=0.7\linewidth]{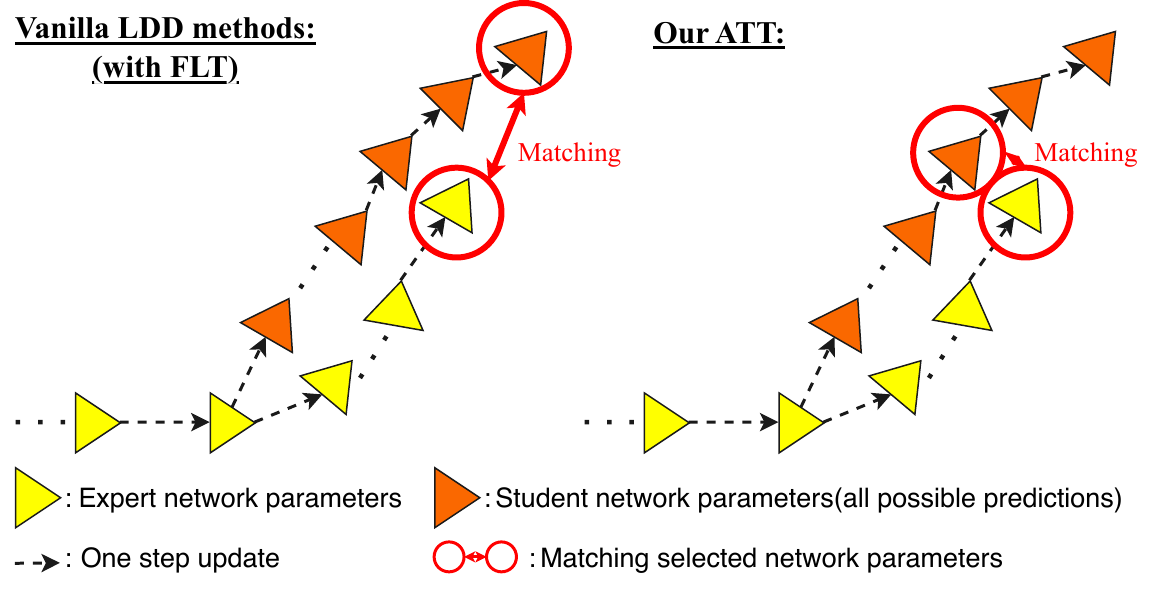}
    \caption{\label{fig:ATT} The figure displays the core idea of our method ATT in comparison to traditional LDD methods employing FTL. Left: The Vanilla LDD bypasses all possible predictions and matches the inaccurate prediction. Right: our method ATT adopts an adaptive approach, aligning predictions with expert network parameters using a minimum distance policy. ATT avoid cases where compressing or stretching trajectories happen, and prevents the accumulation of errors resulting from those cases within the synthetic dataset at each iteration, thus achieves better distillation performances.}
\end{figure*}

\subsection{Automatic Training Trajectories (ATT)\label{subsec:ATT}}
The core idea behind ATT is to dynamically and adaptively select a suitable trajectory length based on the minimum distance policy. And ATT essentially is a simple solution to~\cref{eq:8}. It computes all possible $e_{t}$ and selects the $t$ that gives the minimum $e_{t}$.

\cref{fig:ATT} provides an illustration of the ATT process. The process begins by unrolling a trajectory starting from the expert input $\theta_{i,0}^{*}$ on a synthetic dataset. Within each trajectory, every updated step $t\in\{0,1,\cdots,N_S\}$ conveys a potential prediction $\theta_{i,t}^{\prime}$. We evaluate the squared L2 distance between each prediction and the corresponding expert target with $e_i = \|\theta^{\prime}_{i,t}-\theta_{i,N_T}^{*}\|_2^{2}$.
The step $N_{opt}$ with the minimum distance is selected out of all possible $t$, as depicted in~\cref{eq:8}. Subsequently, back-propagation is performed when matching at the selected prediction $\theta_{i,N_{opt}}^{\prime}$.

We summarize pseudocode for ATT in~\cref{alg}, which provides a step-by-step description of the proposed method. 
By employing this approach, we directly eliminate AMP, and thus reduce matching error.
Further, given the significance of trajectory length in LDD methods, we conduct an ablation study in~\cref{subsec:Ablation Study on Parameters} to investigate its impact on ATT further.

\begin{algorithm}
  \caption{Automatic Training Trajectories \label{alg}}
  {\bf Input:} $N_S$: the number of training steps on $D_S$. \\ 
  \hspace*{11mm}$N_M$: the number of training steps on $D_T$. \\
  \hspace*{11mm}$lr^{\prime}$: learning rate for learning $D_S$
  \begin{algorithmic}[1]
    \STATE{initialize $lr^{\prime}$, $D_S$}
    \WHILE{not converged}
     \STATE{randomly sample a pair of $\theta^{*}_{i,0}$, $\theta^{*}_{i,N_M}$ from expert trajectories}
     \STATE{initialize $\theta^{\prime}_{i,0}=\theta^{*}_{i,0}$, $d_t=[\ ]$}
     \FOR{ $t=0$ to $N_S$}
     \STATE{get $\theta^{\prime}_{i,t+1}$ by a one-step update on $\theta^{\prime}_{i,t}$\\}
     \STATE{calculate $e_{t}$ from \cref{eq:8} and append it to $d_{t}$}
     \ENDFOR
     \STATE{find the index $N_{opt}$, which minimizes $d_{t}$}\\
     \STATE{conduct alternating optimization on~\cref{eq:4} at early step $N_{opt}$ over $lr^{\prime}$ and $D_S$}
     \ENDWHILE
  \end{algorithmic}
  {\bf Output:} learning rate $lr^{\prime}$ and distilled data $D_S$
\end{algorithm}

\section{Experiments} 
In this section, we present an overview of the dataset used, the experimental setups, and the baselines employed in our experiments in~\cref{6.1}. Next, we showcase and compare the performance of our proposed method to the baseline approaches in cross-architecture performance~\cref{subsec:Cross}, and on different datasets~\cref{Exp-SOTA}. 
Furthermore, we perform an ablation study on ATT and highlight its positive impact on boosting stability in~\cref{subsec:Ablation Study on Parameters}. Lastly, we detail ATT's storage and computation requirement in~\cref{subsec:storage and limitaion}.

\subsection{Dataset and Experimental Setup}
\label{6.1}
{\bf CIFAR-10/100~\cite{ds-cifar}:} The dataset comprises 50,000 training images and 10,000 test images, each with a size of 32x32 pixels. CIFAR-10/100 consists of 10 classes and 100 classes respectively. they are colored and natural.\\
{\bf 64x64 Tiny ImageNet~\cite{ds-image}:} The dataset consists of 100,000 training images and 5,000 validation images, all with a size of 64x64 pixels. It encompasses 200 classes and and they are colored and natural.\\
{\bf ImageNet Subsets~\cite{ds-image, imagenet}:} We focus on subset ImageNette (assorted objects), ImageWoof (dog breeds), ImageFruit (fruits), ImageMeow (cats). Each of the subclasses has 10 classes of with the image size of 128x128.\vspace{-5pt}\newline\\
% , ImageSquawk (birds), ImageYellow (Yellow-ish things, e.g., banana and lemon)
{\bf Evaluation Setups:} Following the methodologies established in previous works~\cite{dd-dc, dd-mtt, dd-kip, dd-cafe, dd-dm}, we initiate the distillation process by generating a synthetic dataset using ATT, then assess its performance by training the synthetic dataset with five randomly initialized networks. The test accuracy of these networks is then measured on the original test dataset. To provide a comprehensive assessment, we calculate both the mean and variance of the test accuracy. To ensure a fair comparison, we adopt a simple ConvNet architecture, as designed by~\cite{convnet}, following the precedent set by prior work~\cite{dd-dc, dd-dsa, dd-cafe, dd-mtt}. This ConvNet architecture incorporates convolutional layers~\cite{convnet}, instance normalization~\cite{norm}, activation function RELU~\cite{relu}, and average pooling layers. The normalization layer remains unchanged in our experimental setup to maintain consistency and facilitate a direct comparison of methods.\vspace{-5pt}\newline\\
{\bf Baselines:} In our comparative analysis, we benchmark our research against the first DD method~\cite{dd} and SDD methods including Dataset Condensation (DC)~\cite{dd-dc}, Differentiable Siamese Augmentation (DSA)~\cite{dd-dsa}, Distribution matching (DM)~\cite{dd-dm}, and Aligning Features (CAFE)~\cite{dd-cafe}. Furthermore, we compare DD methods with a meta-gradient computation-solving method: Kernel Inducing Points (KIP)~\cite{dd-kip}. Lastly, we include a comparison with LDD methods, encompassing the first LDD method Matching Training Trajectories (MTT)~\cite{dd-mtt}, Parameter Pruning (PP)~\cite{dd-prunning}, and Flat Trajectory Distillation (FTD)~\cite{dd-mae}. 
To facilitate a fair and meaningful comparison with different methods, our experimental setups deliberately exclude the incorporation of soft labels, network pruning, the adoption of batch normalization, or any additional factorization or parameterization methods. This intentional avoidance ensures that the comparison focuses on the core aspects of the compared methods without introducing confounding factors associated with the mentioned techniques. In this study, we refrain from comparing with instance selection approaches such as random selection, forgetting samples~\cite{cs-forgetting}, or Herding~\cite{cs-herding1}. This decision is based on the understanding that these methods have been significantly surpassed by dataset distillation in many prior works~\cite{dd-dc, dd-dsa, dd-cafe, dd-mtt}. Additionally, we do not include research that concentrates on soft-labels~\cite{dd-sl, dd-sl1, dd-tasla}, data parameterization~\cite{dd-idc, dd-am, dd-slimmable}, data factorization~\cite{dd-dcfactor, dd-haba, dd-prior, dd-gan}, data and label encoding~\cite{dd-am}, selecting representative real samples~\cite{dd-ddrame}, model augmentation~\cite{dd-model-aug}, reduced budget for storage and transmissions with slimmable DD~\cite{dd-slimmable}, approaches with only a few networks~\cite{dd-fewshot}, or with model pool and label learning~\cite{dd-frepo}. The exclusion of these methods is motivated by their orthogonality to our specific research focus, most of them can be implemented in conjunction with our methods as needed.\vspace{-5pt}\newline\\
{\bf Evaluation Details on CIFAR-10/100:}
We adopt the simple 3-layer ConvNet architecture~\cite{convnet} ensure a justifiable comparison with other approaches. To maintain consistency with prior research~\cite{dd-kip, dd-mtt}, we incorporate ZCA whitening and simple data augmentation techniques. Nonetheless, we did not employ Differentiable Siamese Augmentation (DSA)~\cite{dd-dsa} in our baseline method to align identical setups with other baselines. Lastly, to ensure the justification and fairness of our comparison, we adopt a consistent test and learning setup aligned with other LDD methods~\cite{dd-mtt, dd-mae}. This choice aims at providing a fair platform for evaluating the effectiveness and performance of our approach in comparison to existing LDD methodologies.\vspace{-5pt}\newline\\
{\bf Evaluate Details on Tiny ImageNet and ImageNet Subsets:}
Following previous MTT methods~\cite{dd-mtt,dd-mae}, we extend our methodology to larger datasets. Specifically, we utilize the 4-layer ConvNet and 5-layer ConvNet as distillation models for Tiny ImageNet and ImageNet, respectively. To validate the effectiveness of our approach, we conduct experiments on six ImageNet subsets that were previously introduced to dataset distillation by~\cite{dd-mtt}. This ensures that the evaluation is consistent and enables a meaningful assessment of the performance of our approach in comparison to existing methodologies.

\begin{table}[tb]
\setlength{\tabcolsep}{2pt}
\notsotiny
\centering
\caption{The table presents the cross-architecture performance of CIFAR-10 with IPC=10. The network architectures listed in the table remain consistent with those used in previous works~\cite{dd-mtt,dd-mae}, with no modifications made to normalization or pooling layers to ensure fair performance comparisons. We reproduced MTT's result on IPC=50 and FTD's result on IPC=10 for a thorough assessment. The scores are all in percentage. Our method excels significantly in cross-architecture performance under various IPCs.\label{tb:ca}}
    \begin{tabular}{lllllllll} %6 col
    \toprule
    &\multicolumn{2}{c}{ConvNet} & \multicolumn{2}{c}{ResNet18} & \multicolumn{2}{c}{VGG11} & \multicolumn{2}{c}{AlexNet} \\
    \midrule
    IPC&10&50 &10&50 &10&50 &10&50\\
    KIP~\cite{dd-kip}& $47.6\pm0.9$&-& $36.8\pm1.0$ &-&$42.1\pm0.4$&-& $24.4\pm3.9$&- \\
    DSA~\cite{dd-dsa}&$52.1\pm0.4$ &-& $42.8\pm1.0$ &-&$43.2\pm0.5$&-&$35.9\pm1.3$&- \\
    MTT~\cite{dd-mtt}&  $64.3\pm0.7$&$71.6\pm0.2$& $46.4\pm0.6$ &$61.9\pm0.7$&$50.3\pm0.8$&$55.4\pm0.8$&$34.2\pm2.6$&$48.2\pm1.0$\\
    FTD~\cite{dd-mae}&$66.1\pm0.3$&$73.8\pm0.2$&$53.2\pm1.4$&$65.7\pm0.3$&$47.0\pm1.5$&$58.4\pm1.6$&$35.9\pm2.3$&$53.8\pm0.9$\\
    ATT& $\bf 67.7\pm0.6$ &$\bf 74.5\pm0.4$& $\bf 54.5\pm0.9$ &$\bf 66.3\pm1.1$& $\bf 54.2\pm0.8$ & $\bf 61.7\pm0.9$& $\bf 43.6\pm1.4$&$\bf 60.0\pm0.9$\\
    % &-&-&-&-\\
    \bottomrule
    \end{tabular}
\end{table}

\subsection{Cross-Architecture Generalization}
\label{subsec:Cross}
In the evaluation of DD, Cross-Architecture (CA) generalization serves as a crucial metric, gauging the synthetic dataset's capacity to be effectively learned by various architectures. Despite its significance, many prior works struggled to exhibit satisfactory generalization across different architectures. In this subsection, we delve into the examination of ATT in terms of cross-architecture generalization.

We evaluate the CA performance of ATT on ResNet18~\cite{resnet}, VGG11~\cite{vgg}, AlexNet~\cite{alexnet}, and the distillation architecture ConvNet, aligning with architectures studied in prior research. To ensure a fair comparison, we opt for simple network architectures to mitigate the impact of high test performance associated with more powerful structures. In line with closely related works, MTT~\cite{dd-mtt} and FTD~\cite{dd-mae}, which assessed CA performance on different IPCs, we present CA results for our methods in both scenarios. Additionally, we replicate their missing CA evaluations on the other IPC, contributing to a comprehensive and balanced assessment.

The results, as presented in \cref{tb:ca}, showcase the remarkable performance of our proposed method. ATT significantly outperforms previous methods in CA generalization. Specifically, compared to the previous best result with IPC=10, ATT achieves an improvement of 1.3\% on ResNet18, 7.2\% on VGG11, and 7.7\% on AlexNet. Furthermore, for IPC=50, ATT demonstrates improvements of 3.3\% on VGG11 and 6.2\% on AlexNet.

This substantial improvement can be attributed to the mitigation of AMP, as discussed in \cref{subsec:bottleneck}. FTL methods, as revealed in previous discussions, tend to force student trajectories to match with an inappropriate trajectory length, leading to AMP. Consequently, FTL methods exhibit overfitting to seen expert trajectories while sacrificing generalization to unseen trajectories. The CA performance of ATT underscores the significance of addressing the AMP, especially when dealing with unseen trajectories from previously unencountered architectures.

\begin{table}[tb]
\setlength{\tabcolsep}{1.6pt}
\notsotiny
\centering
\caption{\label{tb:SOTA}The table showcases performance comparison to benchmark methods in terms of test accuracy. LD and DD use AlexNet for CIFAR-10, and all the rest are DD methods using the simple ConvNet for the distillation process. IPC: Images Per Class. The numbers in table are presented in percentage. As shown, our method generally outperforms existing methods on various dataset and IPCs.}
    \begin{tabular}{lllllllll} %9 col
         \toprule
         &\multicolumn{3}{c}{CIFAR-10}& \multicolumn{3}{c}{CIFAR-100}& \multicolumn{2}{c}{Tiny ImageNet}\\
         \midrule
         Full dataset&\multicolumn{3}{c}{$84.8\pm0.1$}& \multicolumn{3}{c}{$56.2\pm0.3$}& \multicolumn{2}{c}{$37.6\pm0.4$}\\
         \midrule 
         IPC & 1 & 10 &50& 1 & 10 &50& 1 & 10 \\
         % \cmidrule(r){2-4} \cmidrule(r){5-7} \cmidrule(r){8-9} 
         DD~\cite{dd}&-&$36.8\pm1.2$&-&-&-&-&-&-\\
         DC~\cite{dd-dc}&$28.3\pm0.5$&$44.9\pm0.5$&$53.9\pm0.5$&$12.8\pm0.3$&$25.2\pm0.3$&-&-&-\\
         DSA~\cite{dd-dsa}&$28.8\pm0.7$&$52.1\pm0.5$&$60.6\pm0.5$&$13.9\pm0.3$&$32.3\pm0.3$&$42.8\pm0.4$&-&-\\
         DM~\cite{dd-dm}&$26.0\pm0.8$&$48.9\pm0.6$&$63.0\pm0.4$&$11.4\pm0.3$&$29.7\pm0.3$&$43.6\pm0.4$&$3.9\pm0.2$&$12.9\pm0.4$\\
         CAFE~\cite{dd-cafe}&$ 30.0\pm1.1$&$46.3\pm0.6$&$55.5\pm0.6$&$12.9\pm0.3$&$27.8\pm0.3$&$37.9\pm0.3$&-&-\\
         KIP~\cite{dd-kip}&$\bf 49.9\pm0.2$&$62.7\pm0.3$&$68.6\pm0.2$&$15.7\pm0.2$&$28.3\pm0.1$&-&-&-\\
         % No Frepo
         % FRePo~\cite{dd-frepo}&$46.8\pm0.7$&$65.5\pm0.4$&$71.7\pm0.2$&$\bf 28.7\pm0.1$&$42.5\pm0.2$&$44.3\pm0.2$&$15.4\pm0.3$&$25.4\pm0.2$\\
         MTT~\cite{dd-mtt}&$46.2\pm0.8$&$65.4\pm0.7$&$71.6\pm0.2$&$24.3\pm0.3$&$39.7\pm0.4$&$47.7\pm0.2$&$8.8\pm0.3$&$23.2\pm0.2$\\
         PP~\cite{dd-prunning}&$46.4\pm0.6$&$65.5\pm0.3$&$71.9\pm0.2$&$24.6\pm0.1$&$43.1\pm0.3$&$48.4\pm0.3$&-&-\\
         FTD~\cite{dd-mae}&$46.8\pm0.3$&$66.6\pm0.3$&$ 73.8\pm0.2$&$ 25.2\pm0.2$&$ 43.4\pm0.3$&$50.7\pm0.3$&$ 10.4\pm0.3$&$ 24.5\pm0.2$\\       
         ATT&$\bf 48.3\pm1$ &$\bf 67.7\pm0.6$& $\bf 74.5\pm0.4$& $\bf 26.1\pm0.3$& $\bf 44.2\pm0.5$&$\bf 51.2\pm0.3$&$\bf 11.0\pm0.5$&$\bf 25.8\pm0.4$\\
         % &-&-&-&-&-&-&-&-\\
         \bottomrule
    \end{tabular}
\end{table}
\subsection{Benchmark Comparison}
\label{Exp-SOTA}
Evaluation of distilled architecture performance serves as a crucial metric in dataset distillation, gauging the efficacy of synthetic dataset learning on refined architectures. Our test performance results are detailed in~\cref{tb:SOTA} and~\cref{tb:SOTA2}. ATT, addressing the AMP, aims at improving generality on trajectories not encountered before. Thus, the result additionally reflect how much they suffer from AMP.

Notably, on datasets with substantial image sizes, such as ImageNet with IPC=10, ATT exhibits a significant improvement over the previous best method: 2.2\% on subset Meow, 1.6\% on subset Nette, 1.2\% on subset Woof, and 1\% on subset Fruit. The impact extends to datasets with increased class complexity, exemplified on Tiny ImageNet, where improvements of 1.3\% on IPC=10 and 0.6\% on IPC=1 are observed. CIFAR-10 experiences notable enhancements as well, with improvements of 1.5\% on IPC=1 over FTD, 1.1\% on IPC=10 over previous best result. CIFAR-100 also sees improvements, with up to 0.9\% at different IPCs. While KIP demonstrate commendable performance in IPC=1 on the CIFAR-10 dataset, ATT significantly outperforms them on larger IPCs and more extensive datasets. 

It is crucial to note that, while test performance on distilled architecture is generally significant, it is not as pronounced as those on unseen networks from unfamiliar architectures, as discussed in~\cref{subsec:Cross}. This disparity arises from FTL methods potentially overfitting to pre-buffered expert trajectories, which does not include other network architectures other than the distilled architecture.

\begin{table}
\scriptsize
\centering
\caption{\label{tb:SOTA2}Performance comparison of our method ATT to other baselines on ImageNet subset. All the scores are presented in percentage. As shown, our method demonstrates significant improvement on large dataset ImageNet, especially on IPC=10}
    \setlength{\tabcolsep}{4pt}
    \begin{tabular}{llllll} %12 col
         \toprule
         &&&\multicolumn{2}{c}{ImageNet Subsets}\\
         % \cmidrule(r){3-8}
         Methods & IPC & ImageNette & ImageWoof & ImageFruit & ImageMeow\\
         \midrule
         Full Dataset && $87.4\pm1.0$ & $67.0\pm1.3$ & $63.9\pm2.0$ & $66.7\pm1.1$ \\
         \midrule
         MTT~\cite{dd-mtt}& 1& $47.7\pm0.9$ & $28.6\pm0.8$ & $26.6\pm0.8$ & $30.7\pm1.6$ \\
         & 10 & $63.0\pm1.3$ & $35.8\pm1.8$ & $40.3\pm1.3$ & $40.4\pm2.2$ \\
         FTD~\cite{dd-mae}&1&$52.2\pm1.0$&$30.1\pm1.0$&$29.1\pm0.9$&$33.8\pm1.5$\\
         & 10 &$67.7\pm0.7$&$38.8\pm1.4$&$44.9\pm1.5$&$43.3\pm0.6$\\
         ATT & 1 & $\bf 52.4\pm1.1$ & $\bf 31.2\pm1.8$ & $\bf 30.0\pm0.8$ & $\bf 34.0\pm1.4$ \\
         & 10 & $\bf 69.3\pm0.8$ & $\bf 40.0\pm1.4$ & $\bf 45.9\pm1.2$ & $\bf 45.5\pm1.6$\\
         % ATT+FTD&1&-&-&-&-&-&-\\
         % & 10 &-&-&-&-&-&-\\
         \bottomrule
    \end{tabular}
\end{table}

\begin{figure}[tb]
\centering
\begin{minipage}[b]{0.48\textwidth}
\includegraphics[width=\textwidth]{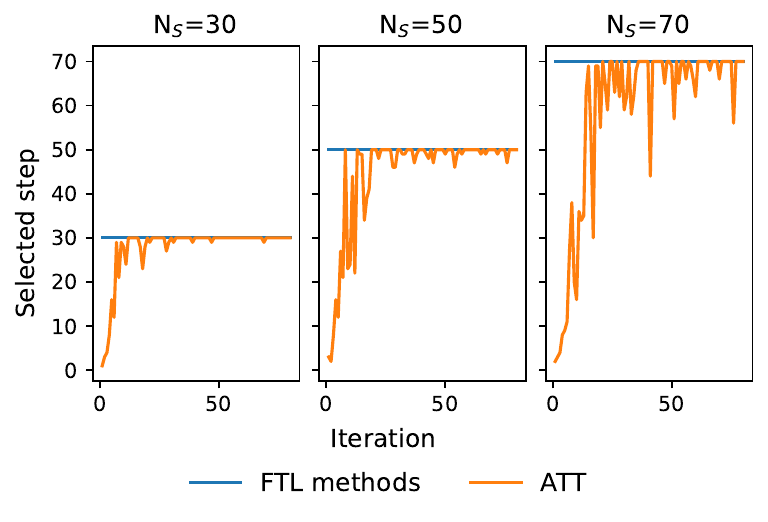}
\end{minipage}
\begin{minipage}[b]{0.48\textwidth}
\includegraphics[width=\textwidth]{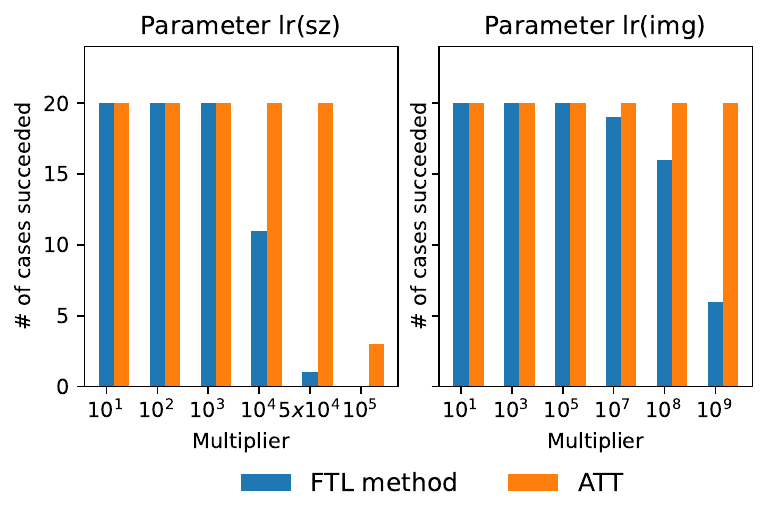}
\end{minipage}
\caption{\label{fig:Ns and parameters}Left: This plot illustrates ATT's step selection during the distillation phase, highlighting its preference for smaller steps initially, contributing to enhanced stability during parameter tuning. Right: The plot demonstrates ATT's superior stability in varying parameters, showcasing the number of successful cases when alter parameter to mutiplier times. Different multipliers are presented as distinct cases.}
\end{figure}

\subsection{Ablation Study on Parameters\label{subsubsec-gs}}
\label{subsec:Ablation Study on Parameters}
\textbf{Trajectory Bounds $N_S$:}
In order to accommodate diverse memory constraints, ATT incorporates the trajectory length parameter $N_S$ to examine the scope of the search for the optimal length. In our experimental analysis, we delve into the behavior of ATT, specifically examining the step selection process during distillation. This investigation focuses on delineating how the optimal step $N_{opt}$ evolves under different trajectory length constraints throughout the distillation process.

Our experiments are conducted on CIFAR-10 with IPC=1, and we emperically choose $N_S$ to be 30, 50, 70, indicating cases with not sufficient steps, close to optimal steps, and too much steps. The results, as depicted on left of ~\cref{fig:Ns and parameters}, highlight the distinction between FTL and ATT. While FTL consistently opts for a fixed trajectory length, ATT tends to dynamically adjust its trajectory length, starting with smaller values at the beginning and subsequently adapting, fluctuating below $N_S$ to minimize errors. The behavior remains with different $N_S$, and we will show more results on various $N_S$ and Iterations in~\cref{appendix:ns and interation}. This adaptive behavior allows ATT to perform short trajectory matching initially instead of enforcing larger trajectory matching.\vspace{-5pt}\newline\\
{\bf ATT and Parameter Variations:} We empirically substantiate ATT's stability toward parameter variation by introducing variations in two sensitive parameters: the learning rate of pixels, denoted as lr(img) and the learning rate of step size, denoted as lr(sz). They are tuning parameters that determine the step size for the synthetic dataset and the trainable learning rate, respectively. We align the initial parameters with the vanilla LDD method, and multiply the examining parameters by factors of multiplier. Specifically, we do not make additional changes to variables other than the one under examination. Each experiment is repeated twenty times, and we document the number of successful cases under each parameter setup. Cases of failure are identified by divergent loss or negative trainable learning rate, while cases with convergent loss are considered successful.

As demonstrated on the right of~\cref{fig:Ns and parameters}, ATT exhibits higher stability under different parameter sets. This is attributed to ATT allowing and selecting a small step size for the student trajectory, facilitating the synthetic data to learn from early trajectories at the beginning, instead of enforcing larger trajectory matching. This advantage holds significant support when undertaking DD on a new dataset.

\subsection{Storage and Computation}
\label{subsec:storage and limitaion}
Similar to other LDD methods such as~\cite{dd-mae, dd-mtt}, computational overhead and storage requirements present challenges. However, LDD methods yield synthetic datasets with superior test performances across various network architectures, benefiting end-users at the cost of increased resources for producers. In our experiments, we primarily use five A100 GPUs, and utilize 100 expert trajectories as the default setting. Assessing the storage requirements for expert trajectories, we find approximately 60MB for each CIFAR expert, and 120MB for each ImageNet expert. We list more details in~\cref{appendix:hw and storage}.

While ATT incurs time in employing distance measurement with L2 norm, it strategically prioritizes matching steps where $t<N_S$, enabling early back-propagation on trajectories and avoiding unnecessary computations at the tail end. This approach serves as a time-efficient compensation for the distance calculation. To illustrate, we conduct experiments on CIFAR-10 with IPC=1, comparing our method to the vanilla LDD approach. The experiments employ identical parameters and hardware setups, repeated ten times, and the average time per iteration is determined. As a result, both the vanilla LDD method and ATT exhibit a similar runtime of 0.8s per iteration.

\section{Conclusions}
In this paper, we delved into long-range matching dataset distillation methods, and reveal that traditional long-range matching methods exhibit persistent limitations and mismatches in aligning trajectories, resulting in the Accumulated Mismatching Problem (AMP). We further reveal AMP's contribution to matching error, and propose ATT to address AMP. Empirical results showcase that ATT not only rectifies the AMP, but also enhances stability when confronted with parameter changes, and the method exhibits superior cross-architecture performance.

As we continue to explore innovative methodologies in dataset distillation, the insights gained from this research pave the way for more effective and adaptable approaches in the field.\newline\\

{\bf Acknowledgment}
This work is funded by Bayerische Forschungsstiftung under the research grants \textit{Von der Edge zur Cloud und zurück: Skalierbare und Adaptive Sensordatenverarbeitung} (AZ-1468-20), and supported by AI systems hosted and operated by the Leibniz-Rechenzentrum (LRZ) der Bayerischen Akademie der Wissenschaften. Further, part of the results have been obtained on systems in the test environment BEAST (Bavarian Energy Architecture \& Software Testbed) at the Leibniz Supercomputing Centre.

\clearpage
\bibliographystyle{splncs04}
\bibliography{main}

\begin{thebibliography}{10}
\providecommand{\url}[1]{\texttt{#1}}
\providecommand{\urlprefix}{URL }
\providecommand{\doi}[1]{https://doi.org/#1}

\bibitem{relu}
Agarap, A.F.: Deep learning using rectified linear units (relu). arXiv preprint arXiv:1803.08375  (2018)

\bibitem{cs-greedy1}
Aljundi, R., Lin, M., Goujaud, B., Bengio, Y.: Gradient based sample selection for online continual learning. Advances in neural information processing systems  \textbf{32} (2019)

\bibitem{cs-u1}
Assadi, S., Bateni, M., Bernstein, A., Mirrokni, V., Stein, C.: Coresets meet edcs: algorithms for matching and vertex cover on massive graphs. In: Proceedings of the Thirtieth Annual ACM-SIAM Symposium on Discrete Algorithms. pp. 1616--1635. SIAM (2019)

\bibitem{cs5}
Bachem, O., Lucic, M., Krause, A.: Practical coreset constructions for machine learning. arXiv preprint arXiv:1703.06476  (2017)

\bibitem{dd-sl1}
Bohdal, O., Yang, Y., Hospedales, T.: Flexible dataset distillation: Learn labels instead of images. arXiv preprint arXiv:2006.08572  (2020)

\bibitem{dd-mtt}
Cazenavette, G., Wang, T., Torralba, A., Efros, A.A., Zhu, J.Y.: Dataset distillation by matching training trajectories. In: Proceedings of the IEEE/CVF Conference on Computer Vision and Pattern Recognition. pp. 4750--4759 (2022)

\bibitem{dd-MTT-fasion}
Cazenavette, G., Wang, T., Torralba, A., Efros, A.A., Zhu, J.Y.: Wearable imagenet: Synthesizing tileable textures via dataset distillation. In: Proceedings of the IEEE/CVF Conference on Computer Vision and Pattern Recognition. pp. 2278--2282 (2022)

\bibitem{dd-prior}
Cazenavette, G., Wang, T., Torralba, A., Efros, A.A., Zhu, J.Y.: Generalizing dataset distillation via deep generative prior. In: Proceedings of the IEEE/CVF Conference on Computer Vision and Pattern Recognition. pp. 3739--3748 (2023)

\bibitem{cs-herding}
Chen, Y., Welling, M.: Parametric herding. In: Proceedings of the Thirteenth International Conference on Artificial Intelligence and Statistics. pp. 97--104. JMLR Workshop and Conference Proceedings (2010)

\bibitem{cs-herding1}
Chen, Y., Welling, M., Smola, A.: Super-samples from kernel herding. arXiv preprint arXiv:1203.3472  (2012)

\bibitem{dd-tasla}
Cui, J., Wang, R., Si, S., Hsieh, C.J.: Scaling up dataset distillation to imagenet-1k with constant memory (2022)

\bibitem{cs-u5}
Dasgupta, A., Drineas, P., Harb, B., Kumar, R., Mahoney, M.W.: Sampling algorithms and coresets for $\backslash$ell\_p regression. SIAM Journal on Computing  \textbf{38}(5),  2060--2078 (2009)

\bibitem{ds-image}
Deng, J., Dong, W., Socher, R., Li, L.J., Li, K., Fei-Fei, L.: Imagenet: A large-scale hierarchical image database. In: 2009 IEEE Conference on Computer Vision and Pattern Recognition. pp. 248--255 (2009). \doi{10.1109/CVPR.2009.5206848}

\bibitem{dd-am}
Deng, Z., Russakovsky, O.: Remember the past: Distilling datasets into addressable memories for neural networks (2022)

\bibitem{dd-privacy}
Dong, T., Zhao, B., Lyu, L.: Privacy for free: How does dataset condensation help privacy? In: International Conference on Machine Learning. pp. 5378--5396. PMLR (2022)

\bibitem{dd-mae}
Du, J., Jiang, Y., Tan, V.T., Zhou, J.T., Li, H.: Minimizing the accumulated trajectory error to improve dataset distillation. arXiv preprint arXiv:2211.11004  (2023)

\bibitem{imagenet}
Fastai: A smaller subset of 10 easily clas- sified classes from imagenet, and a little more french

\bibitem{cs2}
Feldman, D.: Core-sets: Updated survey. Sampling techniques for supervised or unsupervised tasks pp. 23--44 (2020)

\bibitem{convnet}
Gidaris, S., Komodakis, N.: Dynamic few-shot visual learning without forgetting. In: Proceedings of the IEEE conference on computer vision and pattern recognition. pp. 4367--4375 (2018)

\bibitem{dd-datm}
Guo, Z., Wang, K., Cazenavette, G., Li, H., Zhang, K., You, Y.: Towards lossless dataset distillation via difficulty-aligned trajectory matching. arXiv preprint arXiv:2310.05773  (2023)

\bibitem{cs-u2}
Har-Peled, S., Kushal, A.: Smaller coresets for k-median and k-means clustering. In: Proceedings of the twenty-first annual symposium on Computational geometry. pp. 126--134 (2005)

\bibitem{resnet}
He, K., Zhang, X., Ren, S., Sun, J.: Deep residual learning for image recognition (2015)

\bibitem{KD}
Hinton, G., Vinyals, O., Dean, J.: Distilling the knowledge in a neural network. arXiv preprint arXiv:1503.02531  (2015)

\bibitem{dlcv-3}
Kendall, A., Gal, Y.: What uncertainties do we need in bayesian deep learning for computer vision? Advances in neural information processing systems  \textbf{30} (2017)

\bibitem{dd-idc}
Kim, J.H., Kim, J., Oh, S.J., Yun, S., Song, H., Jeong, J., Ha, J.W., Song, H.O.: Dataset condensation via efficient synthetic-data parameterization. In: International Conference on Machine Learning. pp. 11102--11118. PMLR (2022)

\bibitem{dd-apply-medical}
Kiyasseh, D., Zhu, T., Clifton, D.A.: Pcps: Patient cardiac prototypes to probe ai-based medical diagnoses, distill datasets, and retrieve patients. Transactions on Machine Learning Research  (2022)

\bibitem{ds-cifar}
Krizhevsky, A., Hinton, G., et~al.: Learning multiple layers of features from tiny images  (2009)

\bibitem{alexnet}
Krizhevsky, A., Sutskever, I., Hinton, G.E.: Imagenet classification with deep convolutional neural networks. Advances in neural information processing systems  \textbf{25} (2012)

\bibitem{dd-dcfactor}
Lee, H.B., Lee, D.B., Hwang, S.J.: Dataset condensation with latent space knowledge factorization and sharing. arXiv preprint arXiv:2208.10494  (2022)

\bibitem{dd-dcc}
Lee, S., Chun, S., Jung, S., Yun, S., Yoon, S.: Dataset condensation with contrastive signals. In: International Conference on Machine Learning. pp. 12352--12364. PMLR (2022)

\bibitem{dd-prunning}
Li, G., Togo, R., Ogawa, T., Haseyama, M.: Dataset distillation using parameter pruning. arXiv preprint arXiv:2209.14609  (2022)

\bibitem{dd-haba}
Liu, S., Wang, K., Yang, X., Ye, J., Wang, X.: Dataset distillation via factorization (2022)

\bibitem{dd-fewshot}
Liu, S., Wang, X.: Few-shot dataset distillation via translative pre-training. In: Proceedings of the IEEE/CVF International Conference on Computer Vision (ICCV). pp. 18654--18664 (October 2023)

\bibitem{dd-slimmable}
Liu, S., Ye, J., Yu, R., Wang, X.: Slimmable dataset condensation. In: Proceedings of the IEEE/CVF Conference on Computer Vision and Pattern Recognition. pp. 3759--3768 (2023)

\bibitem{dd-ddrame}
Liu, Y., Gu, J., Wang, K., Zhu, Z., Jiang, W., You, Y.: Dream: Efficient dataset distillation by representative matching (2023)

\bibitem{dd-attacks1}
Liu, Y., Li, Z., Backes, M., Shen, Y., Zhang, Y.: Backdoor attacks against dataset distillation. arXiv preprint arXiv:2301.01197  (2023)

\bibitem{dd-attacks2}
Loo, N., Hasani, R., Lechner, M., Rus, D.: Dataset distillation fixes dataset reconstruction attacks. arXiv preprint arXiv:2302.01428  (2023)

\bibitem{dd-cig}
Loo, N., Hasani, R., Lechner, M., Rus, D.: Dataset distillation with convexified implicit gradients (2023)

\bibitem{cs-u3}
Mirrokni, V., Zadimoghaddam, M.: Randomized composable core-sets for distributed submodular maximization. In: Proceedings of the forty-seventh annual ACM symposium on Theory of computing. pp. 153--162 (2015)

\bibitem{dd-kip}
Nguyen, T., Novak, R., Xiao, L., Lee, J.: Dataset distillation with infinitely wide convolutional networks. Advances in Neural Information Processing Systems  \textbf{34},  5186--5198 (2021)

\bibitem{gpuera}
Nickolls, J., Dally, W.J.: The gpu computing era. IEEE micro  \textbf{30}(2),  56--69 (2010)

\bibitem{cs-is}
Olvera-L{\'o}pez, J.A., Carrasco-Ochoa, J.A., Mart{\'\i}nez-Trinidad, J.F., Kittler, J.: A review of instance selection methods. Artificial Intelligence Review  \textbf{34},  133--143 (2010)

\bibitem{dlcv-2}
O’Mahony, N., Campbell, S., Carvalho, A., Harapanahalli, S., Hernandez, G.V., Krpalkova, L., Riordan, D., Walsh, J.: Deep learning vs. traditional computer vision. In: Advances in Computer Vision: Proceedings of the 2019 Computer Vision Conference (CVC), Volume 1 1. pp. 128--144. Springer (2020)

\bibitem{cs-forgetting1}
Paul, M., Ganguli, S., Dziugaite, G.K.: Deep learning on a data diet: Finding important examples early in training. Advances in Neural Information Processing Systems  \textbf{34},  20596--20607 (2021)

\bibitem{dd-fed-4}
Pi, R., Zhang, W., Xie, Y., Gao, J., Wang, X., Kim, S., Chen, Q.: Dynafed: Tackling client data heterogeneity with global dynamics. In: Proceedings of the IEEE/CVF Conference on Computer Vision and Pattern Recognition. pp. 12177--12186 (2023)

\bibitem{dd-continue2}
Sangermano, M., Carta, A., Cossu, A., Bacciu, D.: Sample condensation in online continual learning. In: 2022 International Joint Conference on Neural Networks (IJCNN). pp. 01--08. IEEE (2022)

\bibitem{greenai}
Schwartz, R., Dodge, J., Smith, N.A., Etzioni, O.: Green ai. Communications of the ACM  \textbf{63}(12),  54--63 (2020)

\bibitem{cs1}
Sener, O., Savarese, S.: Active learning for convolutional neural networks: A core-set approach. arXiv preprint arXiv:1708.00489  (2017)

\bibitem{vgg}
Simonyan, K., Zisserman, A.: Very deep convolutional networks for large-scale image recognition. arXiv preprint arXiv:1409.1556  (2014)

\bibitem{dd-fed2}
Song, R., Liu, D., Chen, D.Z., Festag, A., Trinitis, C., Schulz, M., Knoll, A.: Federated learning via decentralized dataset distillation in resource-constrained edge environments. arXiv preprint arXiv:2208.11311  (2022)

\bibitem{dd-sl}
Sucholutsky, I., Schonlau, M.: Soft-label dataset distillation and text dataset distillation. In: 2021 International Joint Conference on Neural Networks (IJCNN). pp.~1--8. IEEE (2021)

\bibitem{cs-forgetting}
Toneva, M., Sordoni, A., Combes, R.T.d., Trischler, A., Bengio, Y., Gordon, G.J.: An empirical study of example forgetting during deep neural network learning. arXiv preprint arXiv:1812.05159  (2018)

\bibitem{cs6}
Tsang, I.W., Kwok, J.T., Cheung, P.M., Cristianini, N.: Core vector machines: Fast svm training on very large data sets. Journal of Machine Learning Research  \textbf{6}(4) (2005)

\bibitem{cs-u4}
Tukan, M., Maalouf, A., Feldman, D.: Coresets for near-convex functions. Advances in Neural Information Processing Systems  \textbf{33},  997--1009 (2020)

\bibitem{norm}
Ulyanov, D., Vedaldi, A., Lempitsky, V.: Instance normalization: The missing ingredient for fast stylization. arXiv preprint arXiv:1607.08022  (2016)

\bibitem{dlcv-1}
Voulodimos, A., Doulamis, N., Doulamis, A., Protopapadakis, E., et~al.: Deep learning for computer vision: A brief review. Computational intelligence and neuroscience  \textbf{2018} (2018)

\bibitem{dd-fed1}
Wang, J., Guo, S., Xie, X., Qi, H.: Protect privacy from gradient leakage attack in federated learning. In: IEEE INFOCOM 2022-IEEE Conference on Computer Communications. pp. 580--589. IEEE (2022)

\bibitem{dd-cafe}
Wang, K., Zhao, B., Peng, X., Zhu, Z., Yang, S., Wang, S., Huang, G., Bilen, H., Wang, X., You, Y.: Cafe: Learning to condense dataset by aligning features. In: Proceedings of the IEEE/CVF Conference on Computer Vision and Pattern Recognition. pp. 12196--12205 (2022)

\bibitem{dd}
Wang, T., Zhu, J.Y., Torralba, A., Efros, A.A.: Dataset distillation. arXiv preprint arXiv:1811.10959  (2018)

\bibitem{dd-continue1}
Wiewel, F., Yang, B.: Condensed composite memory continual learning. In: 2021 International Joint Conference on Neural Networks (IJCNN). pp.~1--8. IEEE (2021)

\bibitem{dd-apply-graphs}
Xu, Z., Chen, Y., Pan, M., Chen, H., Das, M., Yang, H., Tong, H.: Kernel ridge regression-based graph dataset distillation. In: Proceedings of the 29th ACM SIGKDD Conference on Knowledge Discovery and Data Mining. pp. 2850--2861 (2023)

\bibitem{cs3}
Yang, S., Xie, Z., Peng, H., Xu, M., Sun, M., Li, P.: Dataset pruning: Reducing training data by examining generalization influence. arXiv preprint arXiv:2205.09329  (2022)

\bibitem{dd-squeeze}
Yin, Z., Xing, E., Shen, Z.: Squeeze, recover and relabel: Dataset condensation at imagenet scale from a new perspective. Advances in Neural Information Processing Systems  \textbf{36} (2024)

\bibitem{cs-greedy2}
Yoon, J., Madaan, D., Yang, E., Hwang, S.J.: Online coreset selection for rehearsal-based continual learning. arXiv preprint arXiv:2106.01085  (2021)

\bibitem{dd-model-aug}
Zhang, L., Zhang, J., Lei, B., Mukherjee, S., Pan, X., Zhao, B., Ding, C., Li, Y., Xu, D.: Accelerating dataset distillation via model augmentation. In: Proceedings of the IEEE/CVF Conference on Computer Vision and Pattern Recognition. pp. 11950--11959 (2023)

\bibitem{dd-dsa}
Zhao, B., Bilen, H.: Dataset condensation with differentiable siamese augmentation. In: International Conference on Machine Learning. pp. 12674--12685. PMLR (2021)

\bibitem{dd-gan}
Zhao, B., Bilen, H.: Synthesizing informative training samples with gan. arXiv preprint arXiv:2204.07513  (2022)

\bibitem{dd-dm}
Zhao, B., Bilen, H.: Dataset condensation with distribution matching. In: Proceedings of the IEEE/CVF Winter Conference on Applications of Computer Vision. pp. 6514--6523 (2023)

\bibitem{dd-dc}
Zhao, B., Mopuri, K.R., Bilen, H.: Dataset condensation with gradient matching. arXiv preprint arXiv:2006.05929  (2020)

\bibitem{dd-idm}
Zhao, G., Li, G., Qin, Y., Yu, Y.: Improved distribution matching for dataset condensation  (2023)

\bibitem{dd-decentrilise-learning}
Zhmoginov, A., Sandler, M., Miller, N., Kristiansen, G., Vladymyrov, M.: Decentralized learning with multi-headed distillation. In: Proceedings of the IEEE/CVF Conference on Computer Vision and Pattern Recognition. pp. 8053--8063 (2023)

\bibitem{dd-fed-3}
Zhou, Y., Ma, X., Wu, D., Li, X.: Communication-efficient and attack-resistant federated edge learning with dataset distillation. IEEE Transactions on Cloud Computing  (2022)

\bibitem{dd-fed0}
Zhou, Y., Pu, G., Ma, X., Li, X., Wu, D.: Distilled one-shot federated learning. arXiv preprint arXiv:2009.07999  (2020)

\bibitem{dd-frepo}
Zhou, Y., Nezhadarya, E., Ba, J.: Dataset distillation using neural feature regression. arXiv preprint arXiv:2206.00719  (2022)

\end{thebibliography}

\clearpage
\appendix
\section{Appendix}
\subsection{FTL and tolerance $\gamma$}
\label{appendix:gamma}
As previously elucidated in Sec. 3.2, we introduce the parameter $\gamma$ as a measure of tolerance for errors occurring in traditional LDD utilizing FTL. In Sec. 3.2, we illustrate the implementation of $\gamma=2$ in our experiments. In this subsection, we additionally explore the effect of varying $\gamma$ within the range of $[0,5]$, allowing for a maximum deviation of 10\% from the expected outcomes. Notably, previous traditional methods have commonly employed $N_S=50$ as the optimal parameter for experiments conducted on CIFAR-10 with IPC=1. Consequently, we extend our experimental investigations to specifically focus on the parameter $N_S=50$.

\begin{figure}[tb]
\begin{center}
   \includegraphics[width=1\linewidth]{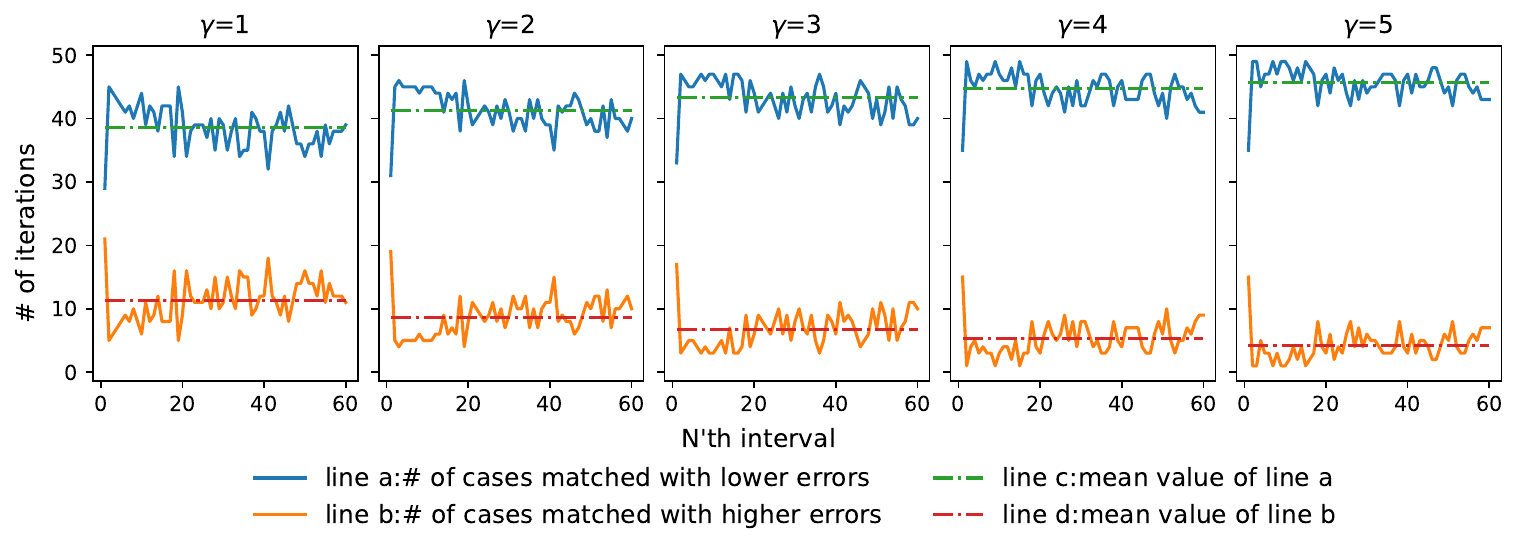}
\end{center}
   \caption{The figure illustrates the impact of the tolerance parameter, $\gamma$, on the performance of FTL. As depicted, the number of iterations exhibits fluctuations under 50, indicating the consistent presence of this phenomenon throughout the entire distillation process. Consequently, it is evident that AMP persists even when a certain level of tolerance for mistakes, as represented by $\gamma$, is applied. This finding emphasizes the resilience of AMP under varying degrees of error tolerance, underscoring the importance of addressing and mitigating this phenomenon in distillation processes.}
\label{fig:append gamma}
\end{figure}

As depicted in~\cref{fig:append gamma}, our findings indicate that, even with different tolerance levels for errors, FTL continues to make incorrect decisions throughout the distillation process. This phenomenon persists despite the introduction of tolerance measures, highlighting the inherent challenges associated with error mitigation in traditional distillation methods.

\subsection{ATT and extensive $N_S$}
\label{appendix:ns and interation}
As presented in Section 4.4, we conduct experiments to examine the trajectory selection behavior of the ATT under different trajectory bounds, specifically $N_S$ values of 30, 50, and 70, using the CIFAR-10 dataset with IPC=1. The results reveal a consistent pattern wherein ATT consistently favored choosing smaller steps in the initial stages of the learning process. We demonstrate more of our results from additional trajectory bounds at 10, 50, 90, 110, and 130.

\begin{figure}[tb]
\begin{center}
   \includegraphics[width=1\linewidth]{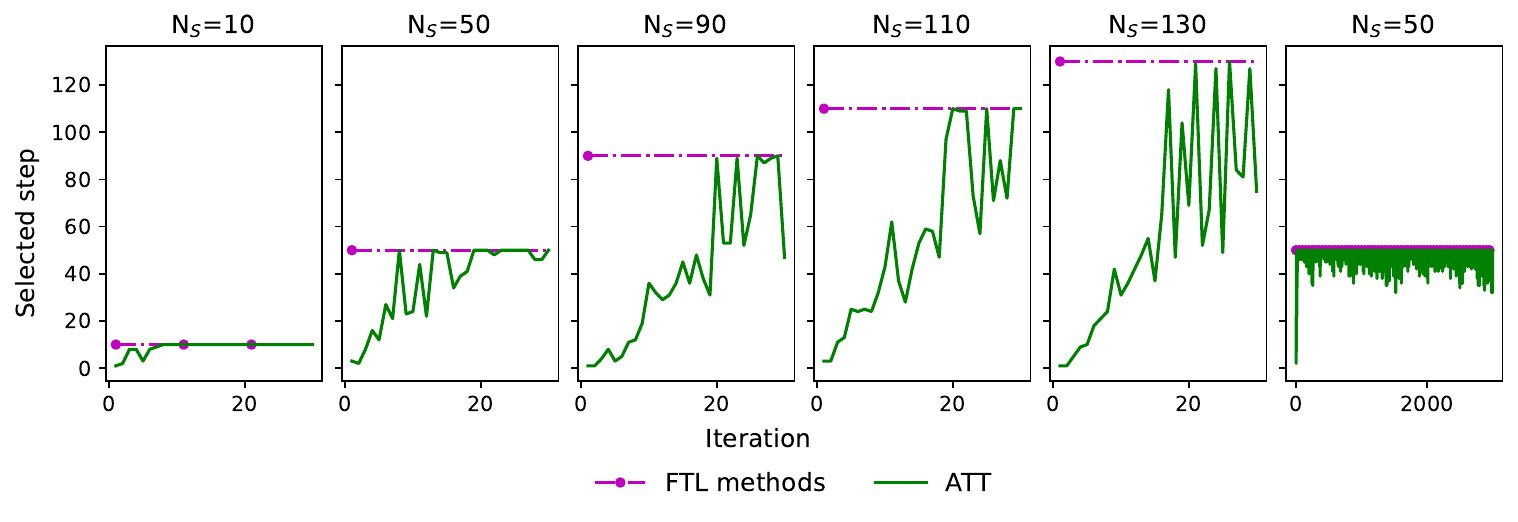}
\end{center}
   \caption{The figure illustrates the influence of extended trajectory bounds, denoted as $N_S$, on the behavior of ATT. As depicted in the first five small plots, regardless of the specific value of $N_S$, ATT consistently exhibits a preference for selecting steps that progress from smaller to larger magnitudes. With the last small plot, we showcase ATT's selection over long distillation, where adjustments maintains. The observed behavior highlights a consistent pattern in the step selection process of ATT, emphasizing its tendency to prioritize learning from smaller steps initially before transitioning to larger ones, irrespective of the trajectory bounds under consideration.}
\label{fig:append ns whole}
\end{figure}

\begin{figure}[tb]
    \centering
    \includegraphics[width=0.5\linewidth]{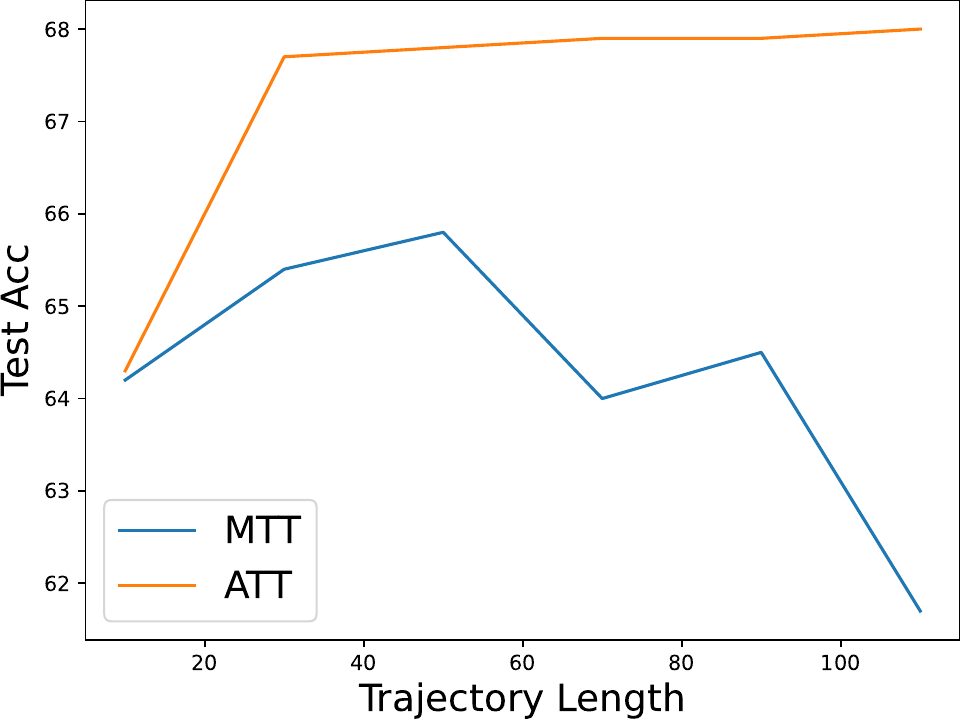}
    \caption{The figure demonstrate variation on ATT's test accuracy with $N_S$. ATT’s performance increasing with $N_S$ and then saturates. The experiments performed on CIFRA-10, and is in comparison with MTT.}
    \label{fig:ATT_NS_Comparison}
\end{figure}

As depicted in the accompanying figure~\cref{fig:append ns whole}, ATT consistently adheres to its established pattern of step selection across a broad range of trajectory bounds, including suitable, extensively small and extensively large values (i.e.10, 50, 90, 110, 130). Notably, ATT demonstrates a predisposition to learn from smaller steps initially, gradually transitioning towards larger steps. This observed tendency supports ATT's stability as discussed in Sec. 4.4. We also demonstrate how ATT's test accuracy varies with $N_S$ in~\cref{fig:ATT_NS_Comparison}, in comparison with MTT. ATT demonstrates stability when varying $N_S$.

\subsection{Random Seeds}
To ensure randomness and non-repetitiveness in our experiments, we maintain the random seeds used in MTT~\cite{dd-mtt} for model initialization. These random seeds are based on time, which means that each time we run the experiments, a different random seed is generated. By using time-based random seeds, we can ensure that the initialized models are random and unique for each initialization and each iteration.

This approach helps to avoid any potential biases or repetitive patterns that may arise from using the same initial model configurations in multiple experiments. By introducing randomness through time-based random seeds, we enhance the reliability and validity of our experimental results by exploring a wider range of possible model initialization.

\subsection{Experimental Parameters}
Important parameters in our method comprise, e.g. the bounds on synthetic steps $N_S$, the number of steps to match on expert trajectories $N_T$, the maximum starting epoch, the learning rate of images $lr(img)$, the learning rate of trainable step size $lr(sc)$, and the initialization of trainable step size $lr$. Further, we maintain the same ZCA normalization as the vanilla LDD method. The listed parameters for each dataset is presented in the table~\cref{tb:hyperparameter}.

\begin{table}[tb]
\setlength{\tabcolsep}{2pt}
\centering
\scriptsize
   \caption{Perparameters for different datasets}
    \begin{tabular}{lllllllllll} %10 col
    \toprule
    Dataset & Model & IPC & $N_S$ & $N_T$ & Max start epoch& lr(img) & lr(sc) & lr& zca\\
    \midrule
    && $1$ &$80$&$4$&$2$&$10^{2}$&$10^{-7}$&$10^{-2}$ &yes\\
    CIFAR-10& ConvNetD3& $10$ & $30$ &$2$ &$20$&$10^4$&$10^{-4}$&$10^{-2}$& yes \\
    && $50$ & $50$ &$2$ &$40$&$10^2$&$10^{-5}$&$10^{-3}$& no \\
    \midrule
    && $1$ &$50$&$4$&$20$&$500$&$10^{-5}$&$10^{-2}$&yes \\
    CIFAR-100& ConvNetD3& $10$ & $20$ &$2$ &$40$&$10^3$&$10^{-5}$&$10^{-2}$&no \\
    && $50$ & $80$ &$2$ &$40$&$10^3$&$10^{-5}$&$10^{-2}$&yes \\
    \midrule
    && $1$ &$30$&$2$&$10$&$10^{4}$&$10^{-4}$&$10^{-2}$&no \\
    Tiny ImageNet& ConvNetD4& $10$ & $20$ &$2$ &$40$&$10^4$&$10^{-6}$&$10^{-3}$&no \\
    \midrule
    && $1$ &$20$&$2$&$10$&$10^{4}$&$5*10^{-8}$&$10^{-2}$&no \\
    ImageNet&ConvNetD5& $10$ & $20$ &$2$ &$10$&$10^4$&$10^{-4}$&$10^{-2}$&no \\
    \bottomrule
    \end{tabular}
  \label{tb:hyperparameter}
\end{table}

\begin{table}[h]
\setlength{\tabcolsep}{3pt}
\centering
\small
\caption{The table presents the number of A100 GPUs and storage used for different experiments. The column "\#Cards" indicates the number of A100 cards we used, and the column "Storage" indicates the storage requirement for storing one expert trajectory}\label{tb:hardwares}
    \begin{tabular}{lllll} %10 col
    \toprule
    Dataset & Model& Storage& IPC & \#Cards\\
    \midrule
    &&& $1$ & $1$\\
    CIFAR-10& ConvNetD3&$\sim 60 MB$& $10$ & $1$\\
    &&& $50$ & $4$\\
    \midrule
    &&& $1$ & $2$\\
    CIFAR-100& ConvNetD3&$\sim 100 MB$& $10$ & $2$\\
    &&& $50$ & $5$\\
    \midrule
    &&& $1$ & $3$\\
    Tiny ImageNet& ConvNetD4&$\sim 170 MB$& $10$ & $5$\\
    \midrule
    &&& $1$ &$1$\\
    ImageNet subsets& ConvNetD5&$\sim 120 MB$& $10$ & $3$\\
    \bottomrule
    \end{tabular}
\end{table}

\subsection{Hardware and Storage Details}
\label{appendix:hw and storage}
We primarily conduct our experiments on A100 GPUs. The largest requirement for GPUs in our experiment was five A100. The following Table \ref{tb:hardwares} presented the number of GPU we used for our experiments in the column "\# Cards", and we also listed the storage requirement for storing one expert trajectory. In our experiments, we use 100 expert trajectories as default for each experts creation.

Long-range Dataset Distillation methods (LDD) generally have high computation overhead and storage requirements. Thus, we keep pre-computed trajectories from the previous method to save memory. Our overheads are roughly similar to the vanilla LDD method.

\subsection{Results Visualization}
\begin{figure}
\begin{center}
   \includegraphics[width=1\linewidth]{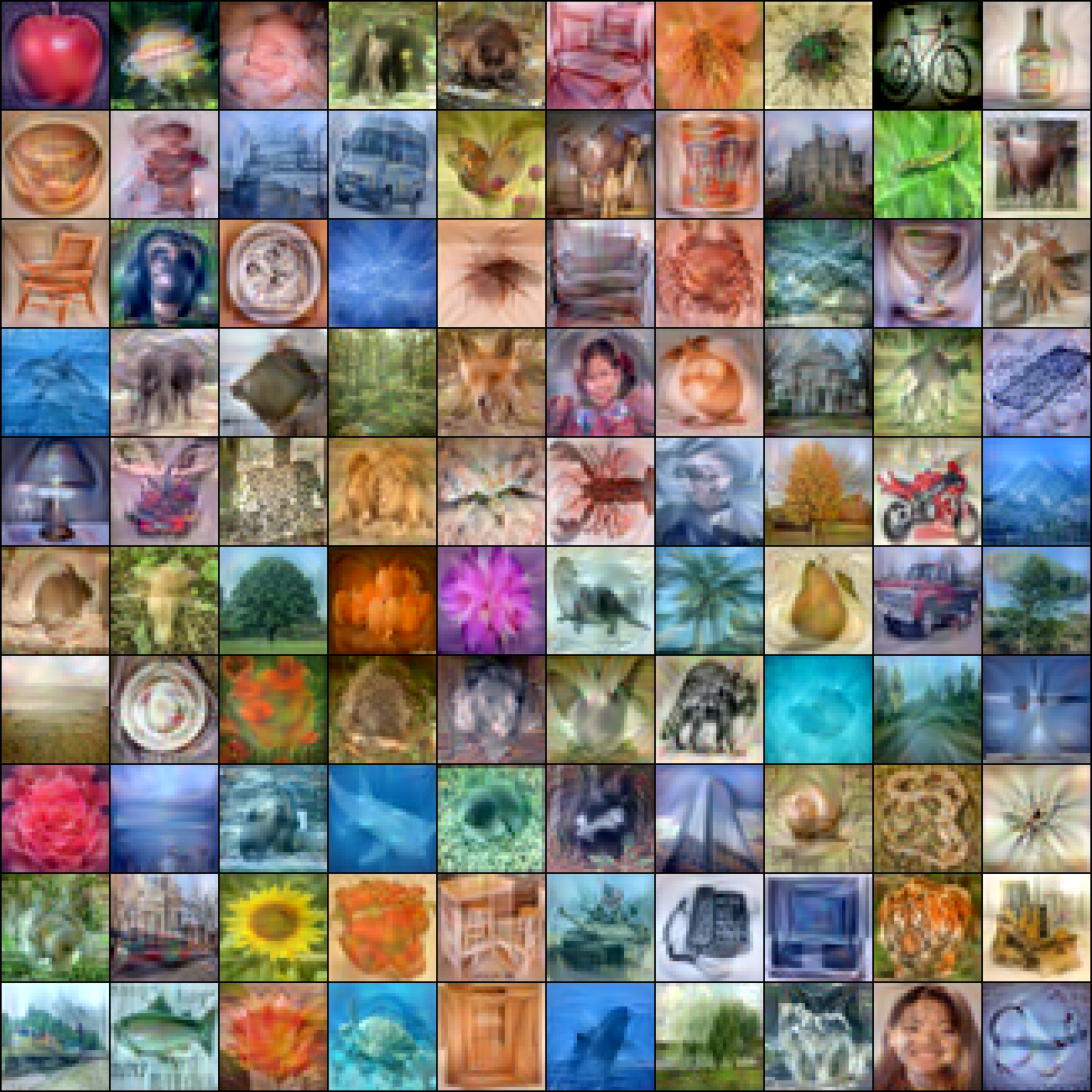}
\end{center} 
   \caption{Results demonstration on CIFAR-100 with IPC=1}
\label{fig:CF100}
\end{figure}

\begin{figure}
\begin{center}
   \includegraphics[width=1\linewidth]{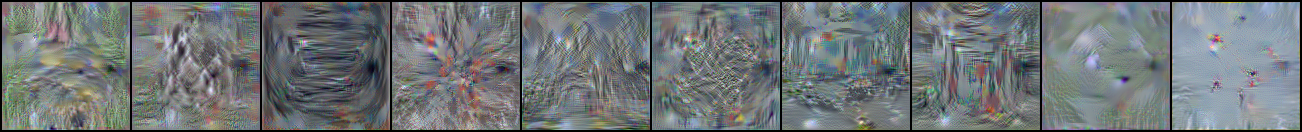}
\end{center} 
   \caption{Results demonstration on ImageNet subset imagenette with IPC=1}
\label{fig:ImageNet}
\end{figure}

\begin{figure}
\begin{center}
   \includegraphics[width=1\linewidth,trim={0 63.9cm 0 0},clip]{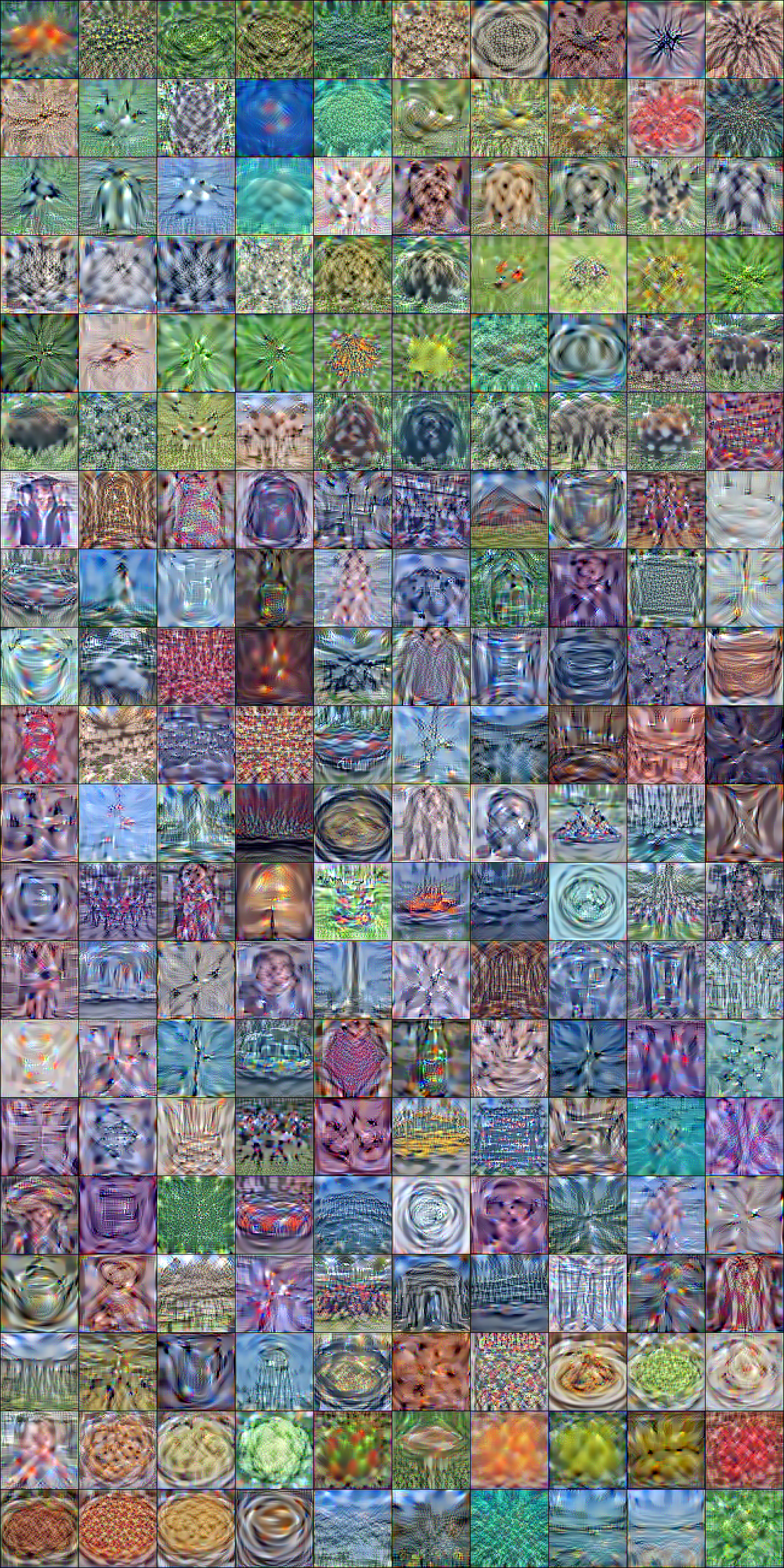}
\end{center} 
   \caption{Partial results demonstration on Tiny ImageNet with IPC=1}
\label{fig:Tiny}
\end{figure}

In this section, we present a selection of results obtained from diverse datasets. Notably, we have previously showcased the results for CIFAR-10 with IPC=1 in the Introduction; therefore, for brevity, these results are not reiterated here. Primarily, we exhibit our findings on CIFAR-100 with IPC=1 in~\cref{fig:CF100}. Subsequently, we demonstrate our results on the ImageNet subset, Imagenette, with IPC=1 in~\cref{fig:ImageNet}. Finally, we illustrate our outcomes on Tiny ImageNet with IPC=1 in~\cref{fig:Tiny}.


\end{document}